\documentclass{article}

\usepackage{arxiv}

\usepackage[utf8]{inputenc} 
\usepackage[T1]{fontenc}    
\usepackage{hyperref}       
\usepackage{url}            
\usepackage{booktabs}       
\usepackage{amsfonts}       
\usepackage{nicefrac}       
\usepackage{microtype}      
\usepackage{lipsum}		
\usepackage{graphicx}
\usepackage[super,sort,comma]{natbib}
\usepackage{doi}

\title{3D Printed Proprioceptive Soft Fluidic Actuators With Graded Porosity}

\author{ \href{}{\hspace{1mm} Nick~Willemstein}\\
	Department of Biomechanical Engineering\\
	University of Twente\\
	Enschede, The Netherlands \\
	\texttt{n.willemstein@utwente.nl} \\
	\And
	{\hspace{1mm}Herman~van der Kooij} \\
	Department of Biomechanical Engineering\\
	University of Twente\\
	Enschede, The Netherlands \\
	\texttt{h.vanderkooij@utwente.nl} \\
	\And
	{\hspace{1mm}Ali~Sadeghi} \\
	Department of Biomechanical Engineering\\
	University of Twente\\
	Enschede, The Netherlands \\
	\texttt{a.sadeghi@utwente.nl} \\}

\renewcommand{\headeright}{}
\renewcommand{\undertitle}{}
\renewcommand{\shorttitle}{\textit{3D Printed Proprioceptive Soft Fluidic Actuators With Graded Porosity}}

\begin{document}
\maketitle

\begin{abstract}
Integration of both actuation and proprioception into the robot body would provide actuation and sensing in a single integrated system. Within this work, a manufacturing approach for such actuators is investigated that relies on 3D printing for fabricating soft-graded porous actuators with piezoresistive sensing and identified models for strain estimation. By 3D printing, a graded porous structure consisting of a conductive thermoplastic elastomer both mechanical programming for actuation and piezoresistive sensing were realized. Whereas identified Wiener-Hammerstein (WH) models estimate the strain by compensating the nonlinear hysteresis of the sensorized actuator. Three actuator types were investigated, namely: a bending actuator, a contractor, and a three DoF bending segment (3DoF). The porosity of the contractors was shown to enable the tailoring of both the stroke and resistance change. Furthermore, the WH models could provide strain estimation with on average high fits (83\%) and low RMS errors (6\%) for all three actuators, which outperformed linear models significantly (76.2/9.4\% fit/RMS error). These results indicate that an integrated manufacturing approach with both 3D printed graded porous structures and system identification can realize sensorized actuators that can be tailored through porosity for both actuation and sensing behavior but also compensate for the nonlinear hysteresis.
\end{abstract}

\section{Introduction}

Soft robotics inherently enable safe cooperation between robots, their environment, and users. This capability is partly realized by their adaptability through their reliance on (mechanically) soft structures. Relying on soft structures instead of rigid mechanical joints is partially inspired by nature, to profit from the versatility seen in biological models such as the elephant trunk and the octopus\cite{trivedi}. 

This inspiration has led to soft actuators and sensors such as those that mimic an octopus' arm\cite{laschi} or skin-inspired sensors (pressure and temperature)\cite{kumaresan}. Furthermore, The flexibility of biological muscles has been the inspiration for many soft robotic actuators such as fluidic soft actuators\cite{LiFOAM,connolly,mosadegh}. These actuators have showcased that a broad range of deformations can be realized such as contraction, twisting, and bending \cite{LiFOAM,connolly,mosadegh}. However, nature also provides design ideas to solve problems such as locomotion strategies based on a caterpillar \cite{lin} or snake \cite{branyan2022curvilinear}, locomotion and grasping based on an octopus \cite{cianchetti2015bioinspired}, and growing to penetrate soil like a plant \cite{sadeghi2014novel}. Another interesting example of this approach is the integration of multiple functions in a single structure, such as the biological muscle. Biological muscles integrate actuation and multiple sensors into a single structure \cite{Ruiz}. This integration includes proprioceptive sensors that are the lengthening of and force exerted by a muscle \cite{Ruiz}. Applying a similar approach to soft actuators would enable the realization of soft systems that can actuate and sense.

Following this strategy of integrating multiple functions, researchers are investigating the embedding of sensors into the body of the robot/actuator. Examples include the usage of conductive fibers and the change in inductance \cite{felt}. Other researchers directly use the change in capacitance/resistance of the actuator itself \cite{Hasel,TCP}. Another approach is to embed smart materials, such as piezo-resistive material, into the body of the actuator \cite{kure,zhou}. 

A popular class of actuators for sensor integration are soft fluidic actuators based on porous structures (such as foam). These foam-like structures are inherently flexible, allow for fluid transport, and can be mechanically programmed\cite{infoam}. Researchers have exploited this property, to realize a broad range of foam actuators, such as contractors, continuum arms, bending actuators, twisting actuators, and pumps \cite{infoam, robertson, murali, murray}. 

In addition, foam-based sensors have been investigated using principles such as optical \cite{meerbeek}, piezocapacitive \cite{bilent,pruvost}, and piezoresistive\cite{murali,foamsense} sensing. Piezocapacitive and -resistive sensing transform the stress/strain to a change in electrical properties by exploiting geometry and/or electrode placement. Such sensors can measure a wide range of deformations such as bending, compression, and shear\cite{foamsense,bilent,murali}. In addition, foam-based sensors can be scaled up by patterning multiple electrodes in a matrix format\cite{cheng}. Lastly, researchers have already demonstrated that sensorized foam can provide proprioceptive data using both piezoresistivity\cite{murali}, optical fibers\cite{meerbeek}, and copper wires (for inductance) \cite{joe2021sensing}. 

Manufacturing of porous/foam-based actuators is often done using commercial foams as a base material. These foams are manually modified using different methods such as laser/manual/wire cutting and assembled using, for instance, gluing\cite{robertson,foamsense, murali,somm2019expanding}. Another popular method is to use lost sugar/salt casting methods\cite{murray, bilent,meerbeek,matsuda2020highly,carneiro2021wearable}, wherein sugar or salt is added during molding/casting, which afterward is removed by washing leading to a porous structure where the salt/sugar was. 

Multiple approaches exist to sensorize these foams, which include placing electrodes on top of either conductive foam for piezoresistive sensors \cite{cheng} or dielectric foam for piezocapactive \cite{bilent} sensors. Another approach is to directly mold a conductive material with lost sugar/salt for piezoresistive sensors \cite{carneiro2021wearable,matsuda2020highly}. Similarly, embedding optical fibers combined with lost sugar/salt has been explored to realize a proprioceptive actuator \cite{meerbeek}. Another common approach is coating/adding a conductive liquid to commercial foam \cite{foamsense, murali,somm2019expanding}, which resulted in proprioceptive actuators \cite{murali,somm2019expanding}. However, to fully exploit the potential of foam-like structures, which includes enabling fluid transport, and lightweight, and mechanical programmability, new integrated fabrication processes need to be developed that can incorporate: a stiffness gradient for deformation programming, smart materials for sensing, and complex geometries. Such capabilities can be attained with 3D printing, which has been shown to enable complex geometries, fabricate multi-material structures, and use soft materials. 

There are multiple approaches to 3D printing porous structures. An approach to 3D print foam-like structures is to add a porogen such as ammonium bicarbonate that degrades to a gas creating pores in the structure \cite{yirmibesoglu2021multi}. Alternatively, the behavior of the material during deposition itself could be exploited, such as liquid rope coiling. The liquid rope coiling effect is the coiling seen when dropping, for instance, honey from a height above a surface \cite{ribe2012}, which is also observed in 3D printing \cite{lipton}. The coiling pattern can be exploited systematically through our previously developed InFoam method \cite{infoam} to 3D print porous structures. Combining this control over the coiling pattern with normal printing allows for 3D-printed structures with a user-defined porosity gradient. Control over the porosity gradient enables mechanical programming, as changes in porosity can lead to stiffness change of more than one order of magnitude \cite{infoam, lipton}). Such a large change in stiffness can be exploited as a stiffness gradient to program the deformation for realizing bending, twisting, and contracting actuators \cite{infoam}. Furthermore, we showed that the performance of bending actuators could be tailored through control over the porosity \cite{infoam}. Therefore, controlling the porosity can be used for mechanical programming. However, these actuators lacked sensing capabilities. Integration of sensors in these actuators would widen the scope of the InFoam method to enable both mechanical programming and sensorization in a simple manufacturing approach.

Within this work, we explore the printing of sensorized soft actuators using the InFoam method by using a conductive thermoplastic elastomer (cTPE). The usage of a cTPE filled with carbon black particles allows the InFoam method to incorporate piezoresistive sensing in the printed structure. The combination of piezoresistive sensing and a porosity gradient allows for the 3D printing of sensorized soft actuators, which both deform and change their resistance due to collapsing pores. Furthermore, the InFoam method's control over the porosity gradient will be shown to enable both the programming of the actuator's mechanical behavior and the strain-resistance change of a contracting actuator. In addition, we show that combining our 3D-printed sensorized actuators with system identification allows us to compensate for the viscoelastic behavior of the piezoresistive sensors to enable strain estimation. Thereby providing a single package that comprises a manufacturing process and a data-driven approach that can estimate the deformation of the printed sensorized actuator by feeding raw data to identified Wiener-Hammerstein models. Within this work, we show this combination of 3D printing and system identification for three actuators, namely: a bending actuator, a contracting actuator, and three degrees of freedom (3DoF) bending segment.

\section{Results}

\subsection{Manufacturing of Soft Sensorized Vacuum Actuators with Deformation Estimation}

\begin{figure*}[h!] 
\centering
\includegraphics[width=0.8\textwidth]{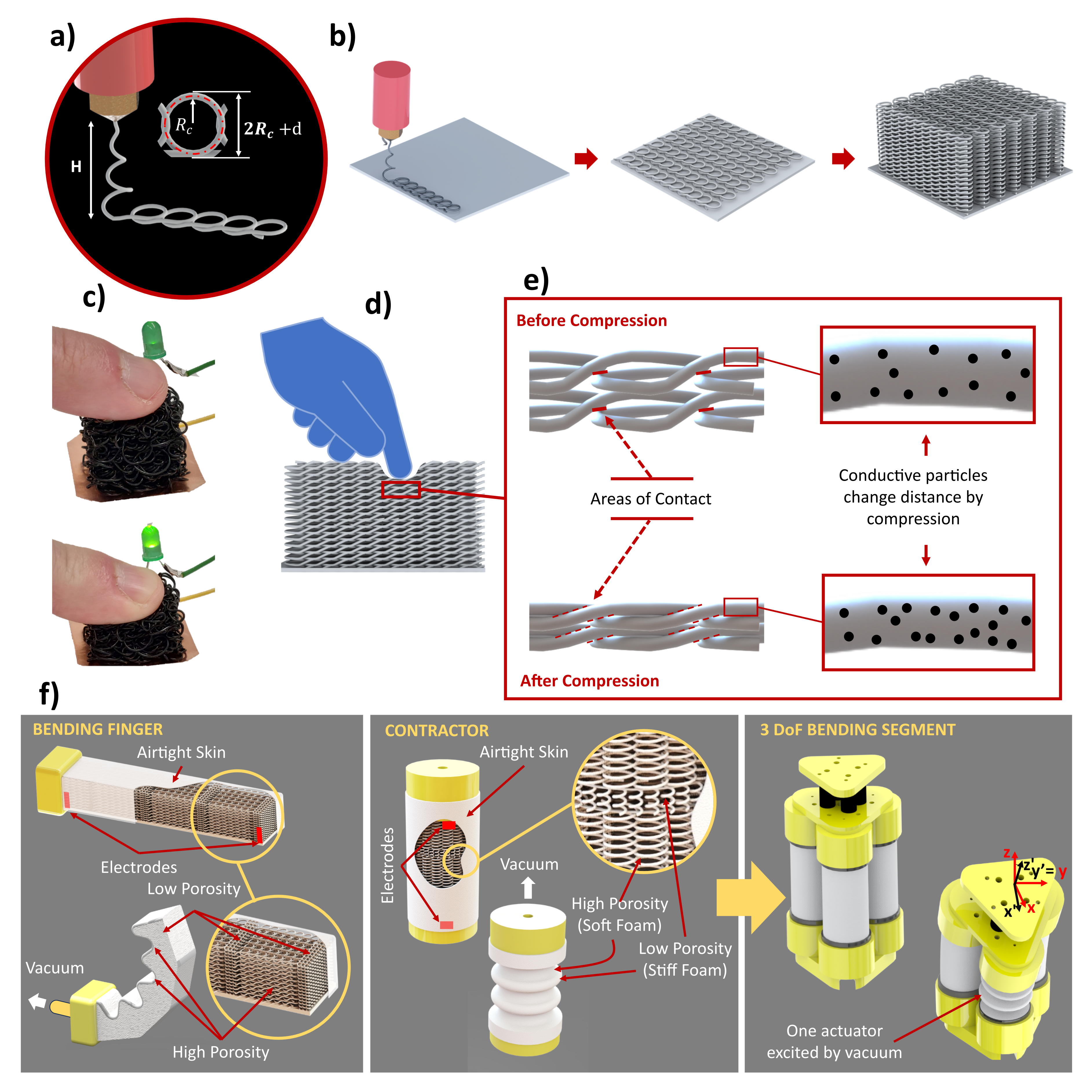}
\caption{a, b) The InFoam methods parameters based on liquid rope coiling (a) and deposition strategy to construct a 3D porous structure (b), c) Resistance change due to the piezoresistive effect by manual compression of a 3D printed porous sensor, 
d, e) Macroscale compression (d) and the piezoresistive effect due to macroscopic (area of contact) and microscale (particle distance) changes (e)  f) Exploiting the porosity gradient through the InFoam method to realize three types of actuators (bending, contraction, and three degrees of freedom (3DoF) bending segment).}
\label{fig:gradact}
\end{figure*}

Soft fluidic actuators based on graded porosity exploit both the ability of porosity to allow for fluid transport but also the significant changes in stiffness due to porosity gradients for programming the deformation. To realize these soft actuators based on graded porosity a fabrication method that can translate a user-defined porosity gradient to a 3D-printed structure is required. Within this work, we use our previously developed InFoam method\cite{infoam} for this purpose. The InFoam method can fabricate porous structures with a porosity gradient by depositing coils of different sizes by exploiting the liquid rope coiling effect (Figure \ref{fig:gradact}(a)) The inherent design freedom of 3D printing allows the InFoam method to fabricate structures with user-defined porosity gradients. The coiling radius $R_c$ and coiling density $N$ define the coiling pattern, which is dependent on the material, machine, and process parameters. During extrusion, the coils are stacked on top of each other, which leads to different angles based on the distance between individual coils. This geometrical property is normalized by the coiling density $N$ \cite{infoam}, which is defined by the number of coils within the outer coil diameter of a single coil. At a fixed temperature $R_c$ and $N$ are primarily determined by the height $H$ and the ratio of extruded amount and distance traveled. The relation between these parameters and the machine was determined through the method described in \cite{infoam} (See Figure S1 in Supporting Information). Based on these results the InFoam method generates the GCode through a custom MATLAB (The Mathworks, Inc., USA) script, which leads to the process visualized in Figure \ref{fig:gradact}(b). This script computes the required extrusion and movement speed and heights based on the coiling pattern geometry to print the 3D graded porous structure \cite{infoam}. 

The graded porosity achieved by the InFoam method can be combined with a (carbon black-filled) conductive thermoplastic elastomer (cTPE) to integrate sensing capabilities into the structure. The cTPE enables sensing due to the change in resistance between points in the structure due to the deformation history \cite{paredes, foamsense,murali}.  The resistance changes can be very significant ($>$80\%, as will be shown in the next sections and Supporting Movie 1) as seen by the brightness change in Figure \ref{fig:gradact}(c). An advantage of these sensors is that resistance measurements can be done using simple electronics. Specifically, the change of resistance is often measured, which is defined as
\begin{equation}
    \Delta{R}(t) = \frac{R(t)-R_0}{R_0}\cdot 100\%
\end{equation}
Wherein the variables $\Delta R$, $R$, $R_0$ are the relative change of resistance (\%), and the instantaneous and initial (at no strain/stress) resistance ($\Omega$). The manufacturing of these soft sensorized actuators is done in two steps. Firstly, print using both conventional 3D printing and our InFoam Method. Afterwards, the porous structure is assembled to finalize the sensorized actuator with a skin. The fabricated sensorized actuators, however, require model-based compensation to estimate the strain based on the resistance change. Because the 3D-printed piezoresistive foam-like structure does not have a simple linear relationship with the deformation, as they exhibit hysteresis.

This hysteresis is expected to be due to both macroscopic and microscopic effects. On the macro-level resistance changes in the porous cTPE structure are due to a change in surface contact of the coils (Figure \ref{fig:gradact}(d) and (e)), which is affected by both the amount of deformation and the viscoelastic properties of the cTPE itself. However, on the micro-level, the coils deform leading to a change in conductive particle distance and density which also results in a resistance change of the overall structure/actuator. On the micro level, the carbon-black particles have a degree of randomness, as the exact configuration will change between each actuation cycle.  By applying stress these particles can be forced closer together leading to a change in resistance. Due to the change in inter-particle distance and density a change of resistance will happen\cite{paredes,kalantari2011new}. 
 
However, the particles are in the end moved by the elastomeric matrix that surrounds them. As the actuator's pores collapse, macro-level effects will play a major role. Especially, as unlike the compression of a bulk elastomer \cite{paredes,kalantari2011new} the coils of the porous structure collapse and are pressed on top of each, which will decrease the contact resistance between coils (see Figure \ref{fig:gradact}(d)). On this macro-level the cTPE foam-like structure itself exhibits both viscoelastic and nonlinear collapse properties \cite{Goga,paredes}. The former will lead to hysteresis as the behavior during loading and unloading will be different and strain-rate dependent. Whereas the latter will decrease the sensitivity for higher compressive strains. This change can be explained as during initial buckling the increase in contact area will decrease the resistance significantly as all the coils will come into contact with each other. In contrast, the coils are pressed more firmly after the initial buckling, but this change is significantly less, as the contact area increases much slower. 

Thus, the cTPE porous structure provides piezoresistive sensing capabilities. The InFoam method can fabricate this porous structure but also mechanically program it by the porosity gradient \cite{infoam}. This mechanical programming is possible due to the porosity gradient's significant effect on the modulus (over two orders of magnitude). This large difference in stiffness can be exploited to use the porosity gradient to realize a bending actuator and contracting actuators (Figure \ref{fig:gradact}(f)). Moreover, by combining multiple contractors in parallel, a three-degree of freedom bending segment can be created as well (Figure \ref{fig:gradact}(f)). Combining this mechanical programming capability with a soft conductive material would enable actuators with programmable deformation and integrated sensing. These three actuators (bending actuator, contractor, and 3DoF bending segment) were investigated in this work for strain estimation. These three actuator types were selected as these are widely used modules in many soft robotic systems such as grippers, artificial muscles, and continuum arms. In addition, they include both single input and output systems (bending actuator and contractor) and a multiple inputs and outputs system (the 3DoF bending segment).

\begin{figure*}[h!] 
\centering
\includegraphics[width=0.5\textwidth]{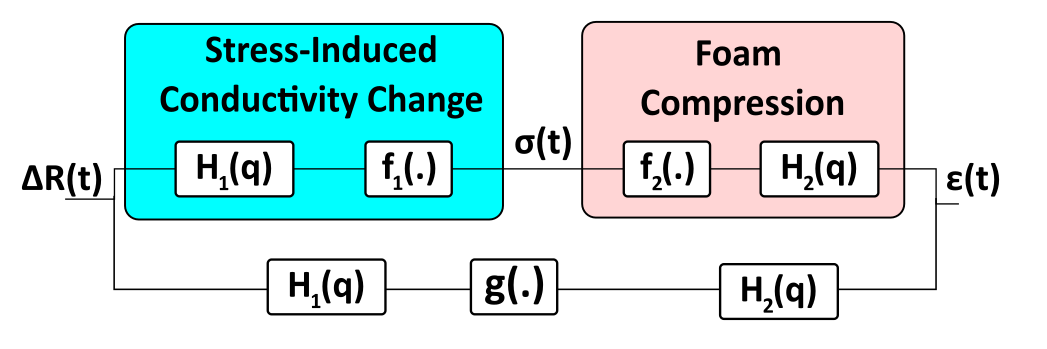}
\caption{The Wiener-Hammerstein model used for strain estimation.}
\label{fig:WHmodel}
\end{figure*}

Although these actuators can be fabricated from a cTPE material to sensorize them, the step from resistance change to strain is not straightforward. A key challenge herein is that the piezoresistive behavior of our cTPE structure exhibits nonlinearities and (viscoelastic) hysteresis. To overcome this challenge, a system identification approach is used to identify a model that can estimate the strain of the actuator based on the resistance change over time. Specifically, the usage of a Wiener-Hammerstein model (WH model) was investigated (Figure \ref{fig:WHmodel}). The WH model consists of two linear systems and a static nonlinearity in the middle. The combination of linear and nonlinear functions allows the WH model to capture the inherent nonlinear couplings between resistance change and strain but also take the deformation history into account (to compensate for hysteresis). A similar model (Wiener) has shown good results for force sensing resistors \cite{Saadeh} but has, to our knowledge, not been applied to strain sensing of soft actuators. Furthermore, we investigated the use of the Wiener-Hammerstein form, which was not (to our knowledge) investigated before for strain estimation. 

The WH model structure should mimic the structure of the underlying physics. Firstly, the left part models the stress-driven resistance change\cite{paredes}. Furthermore, it has been shown that the stress $\sigma(t)$ is captured well by a Wiener model\cite{Saadeh} 
\begin{equation}
    \sigma(t) = f_1\left( \left(H_1(q \right) \Delta R(t)\right)
\end{equation}
The functions $f_1$ and $H_1(q)$ represent a nonlinear mapping and linear transfer function, respectively. \\
Whereas $q$ represents the time-shift operator such that $x(t)q^{-n}=x(t-n)$ (for discrete-time systems). The first stage of the model relates the resistance change and stress. The stress and strain are also nonlinearly related for porous structures\cite{infoam, Liu, Goga}. One possible approximation is through a nonlinear viscoelastic model to relate the strain to stress\cite{Goga}. Within this work, a Hammerstein (HS) model is used as an approximation of the strain $\epsilon(t)$
\begin{equation}
    \epsilon(t) = H_2(q)f_2\left(\sigma (t)) \right)  
\end{equation}
Within this equation, the functions $f_2$ and $H_2(q)$ represent a nonlinear mapping and a viscoelastic model (transfer function), respectively. Combining these two equations in series leads to the following equation
\begin{equation} \label{eq:model}
    \epsilon(t) = H_2(q)\left(g\left(H_1(q)\Delta R(t)) \right) \right)
\end{equation}
Within this equation, the two-stage static nonlinearity (i.e. $f_1(f_2(.))$) is reduced to a single one $g(.)$. This function $g(.)$ approximates the combined nonlinear functions. Within this work, $g(.)$ was implemented as a piecewise linear function. This function type was selected and has been used before in \cite{Saadeh} for stress estimation and promising results (NRMSE fits) during (initial) manual estimation of the models. Lastly, it should be noted that the model in Equation \ref{eq:model} can be reduced to an HS or Wiener model by either removing $H_2(q)$ or $H_1(q)$, respectively.

\subsection{Curvature Sensing \& Reconstruction}

To investigate the capability of curvature sensing, a sensorized vacuum-based bending actuator was 3D printed and characterized. The actuator was fabricated using different percentages of porosity as seen in Figure \ref{fig:gradact}(f)). Firstly, a layer of bulk material (zero porosity) with a thickness of 1 mm was printed at the bottom. On top of this bulky layer, four sections with high porosity of 84\% (75x15x10 mm$^3$ length x width x height) together with four spacers of low porosity (below 5\%) were printed. The printed structures were then packed in a 0.4 mm thick heat-sealed styrene-ethylene-butylene-styrene (SEBS) sleeve to finalize the actuator. The SEBS sleeve could deform and collapse the high porosity sections when negative pressure was applied. Whereas the bulk layer acted as a (relatively) inextensible layer, which provided an asymmetry to the structure, which led to bending. The low porosity spacers increase the achievable bending angles while still allowing the passage of air through the actuator. Electrodes (copper wires) were then connected to the top left and bottom right of the actuator to measure the resistance. 
The bending actuator was tested by putting the bending actuator in a holder printed from polylactic acid (PLA). The actuator was excited by multiple levels of vacuum (gauge pressures of -20,-40,-60 kPa) with three on/off (step) cycles and a ``gradual increase" signal (0 to -60 and back to 0 kPa with intermediate stops). During these experiments, the resistance change was measured using an Arduino Uno (Arduino AG, Italy). Lastly, to track the curvature a webcam was used to capture images of the actuator, which was subsequently processed to acquire the deformation. The setup is shown in Figure S2 in the Supporting Information, which includes the definition of the curvature.

\begin{figure}[h!]
\centering
\includegraphics[width=0.5\textwidth]{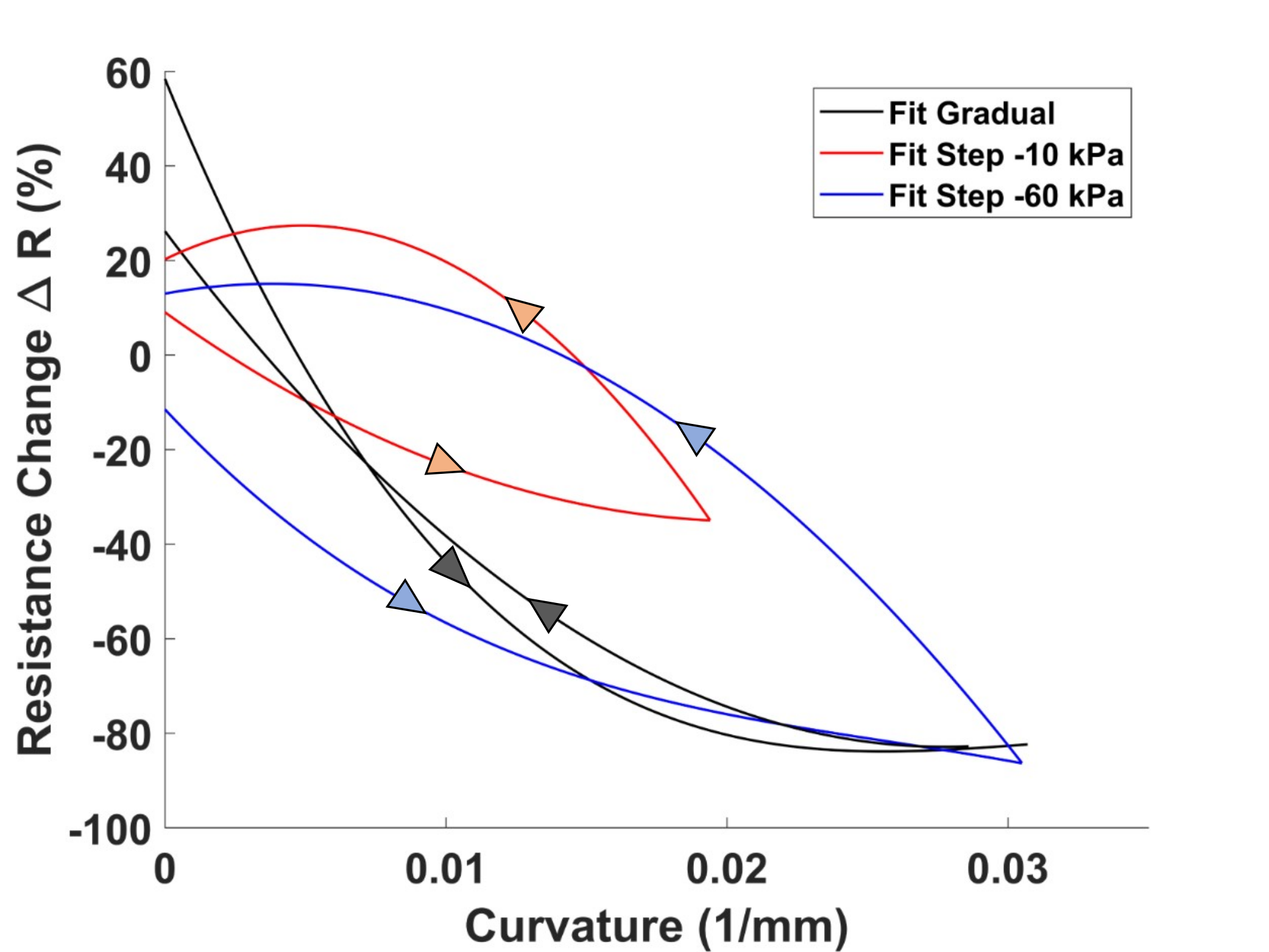}
\caption{The hysteresis curve of resistance change versus curvature for two steps in pressure (-20 and -60 kPa) and the ``gradual increase" dataset. These are fitted curves in both directions, for the hysteresis plot including the raw data points the reader is referred to the Figure S2 in the Supporting Information}
\label{fig:bend}
\end{figure}

The curvature versus change in resistance is shown in Figure \ref{fig:bend} for three different pressure series. It can be observed that the resistance decreases with increasing pressure and curvature in all datasets. This behavior is to be expected as when the pores collapse the contact area increases. It can be observed that both the gradual increase and -60 kPa step end at similar resistance changes and curvature. This behavior implies that steady-state behavior is not dependent on history. However, for all three a clear hysteretic behavior can be observed in the transient response (i.e. before steady-state. Furthermore, there is an overshoot when relaxing leading to a higher resistance than initial. This discrepancy decays before the next cycle similar in behavior to viscoelastic relaxation.

When we fit polynomial functions (third or second order) to approximate the hysteresis curve, a clear hysteresis can be observed in all three scenarios. This behavior means the relationship between resistance change and curvature is not a simple curve and depends on the rate of change. The latter can be observed from the area of hysteresis that seems independent of the magnitude of the vacuum pressure as the -10 kPa-step seems to have more hysteresis than the ``gradual increase"-dataset even though the latter reaches -60 kPa. The difference in pressure increase (i.e. gradual versus step) implies that the rate plays a significant factor. Such behavior is expected, as existing models of carbon black-based piezo-resistive sensors include a damping component \cite{paredes}, which would lead to rate-dependent behavior. This rate dependence implies that the change in resistance behaves viscoelastically (in line with our hypothesis). Furthermore, the rate dependence means that a proper model of these sensors will need to take the history into account. Lastly, there also seems to be a nonlinear relationship between resistance change and curvature, which will need to be considered as well by such a model.

To compensate for the observed hysteresis and nonlinearity a data-driven approach is used by identifying a set of models and evaluating their performance. Four types of models were identified, namely: linear, HS, Wiener, and Wiener-Hammerstein (WH). The latter was acquired by first identifying a Wiener/ linear model and then using that as a filter to refine the output by adding a linear/HS model in series. The data was separated into an identification (``gradual increase" and the -60/-10 kPa) and validation set (-20 and -40 kPa).

The models were evaluated using several metrics to evaluate their estimation capability. Firstly, the  normalized root mean square (NRMSE) fits of the model were evaluated through MATLAB's compare-function as:
\begin{equation}
    \textrm{NRMSE fit} = 100\left(1-\frac{||y-\hat{y}||}{||y-\bar{y}||}\right)
\end{equation}
In this equation, the NRMSE fit is computed using the two norms of the measured $y$ and predicted $\hat{y}$ output divided by the difference between the estimated value and average output $\bar{y}$. The NRMSE fit was used to select the best model, as the model that performed best on average on the identification and validation datasets was taken as the best model, Similarly, the RMS error was computed and scaled by the maximum output (and the number of data points $N$) in the dataset, as
\begin{equation}
    \textrm{RMS} =\frac{100}{\max(y)} \sqrt{\frac{\sum_{i=1}^{N} (y(i)-\hat{y}(i))^2}{N}}
\end{equation}

\begin{figure*}[h] 
\centering
\includegraphics[width=0.95\textwidth]{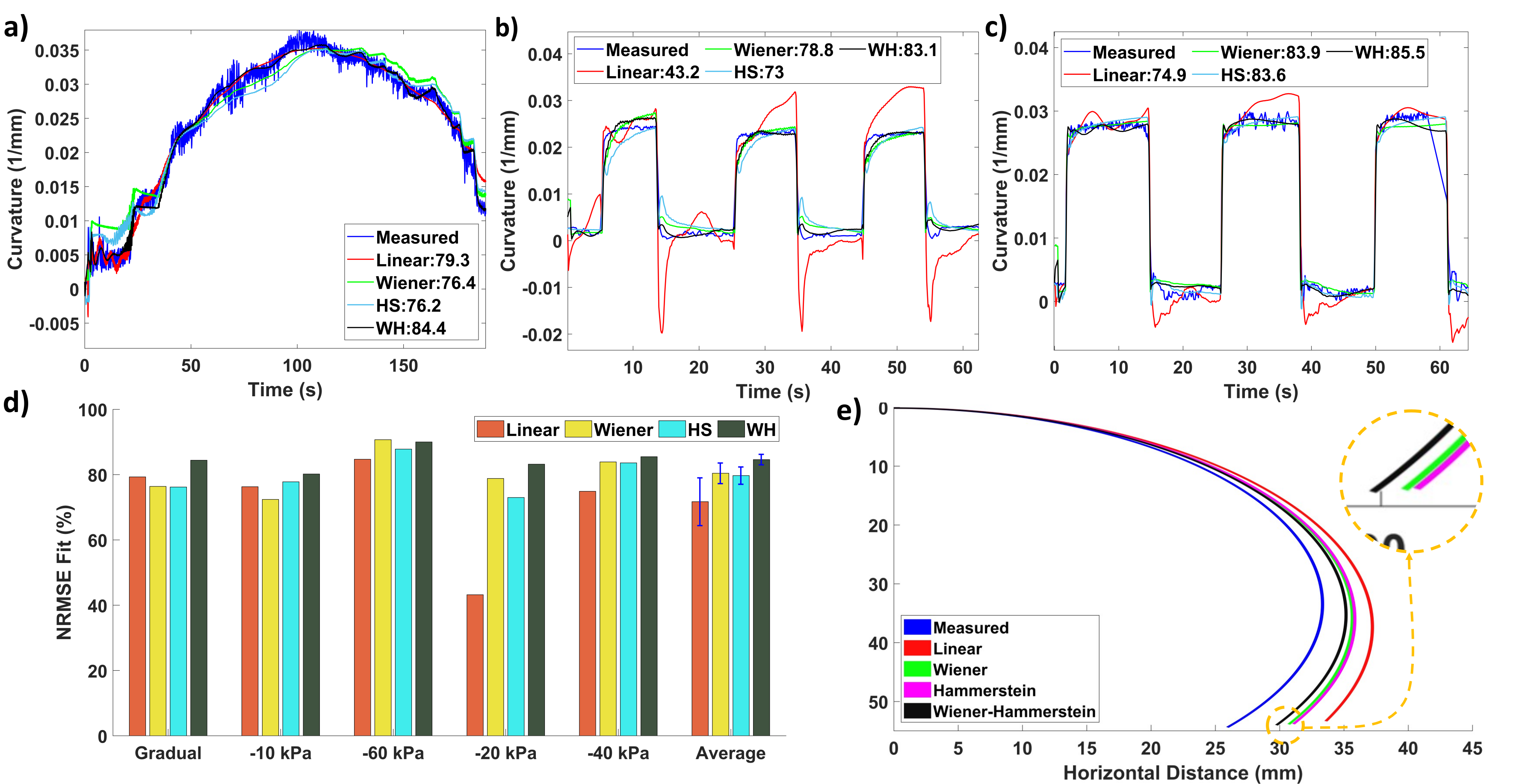}
\caption{a-c) The bending actuator’s measured and predicted curvature by the identified models (incl. NRMSE fit in legend) for (a) the ``gradual increase"-dataset (identification) and both the validation datasets (b) -20 and (c) -40 kPa. d) A bar graph with the NRMSE fits of all identified models, and e) the RMS error visualized at the maximum curvature. Abbreviations: WH and HS mean Wiener-Hammerstein and Hammerstein models, respectively.}
\label{fig:curvaturereconstruction}
\end{figure*}

By scaling the RMS error by the maximum deformation in the dataset a comparison between datasets can be made. This evaluation approach also means that the RMS error (in percentage) can be scaled by the maximum deformation to estimate the RMS error in deformation, as these have the same unit. 
The objective for identification was set to estimate the curvature, which was defined as the inverse of radius $r$ of the circle that the bending actuator made (see Figure S2 in the Supporting Information as well). A set of models were estimated based on the identification datasets (one for each model type). These models were evaluated based on their average NRMSE fit and the best one was kept. The predictions of these identified models are shown in Figure \ref{fig:curvaturereconstruction}(a-c), which includes a set from identification and two validation sets. All models seem capable of capturing the dynamics of the ``gradual increase" dataset quite well. In line with our hypothesis, the WH model outperforms the other three. Whereas the Wiener and HS have a lower fit than the linear model only for the ``gradual increase" dataset. However, looking at how well they approximate the curve the linear model just averages whereas the nonlinear models try to capture the faster dynamics and nonlinearities. This discrepancy is especially apparent in the other two datasets (i.e. the validation). The fits for 20 kPa clearly show that the linear model has significant problems with capturing the dynamics properly leading to overshooting. However, this difference is significantly less apparent for the -40 kPa dataset whereas the HS and Wiener model is similar to the WH models' performance, which indicates that the nonlinear models generalize better (see also Supporting movie 2). 
It is expected that this error discrepancy is much less for the 40 kPa dataset as foam-like structures collapse and have distinct regimes of change\cite{pruvost}. These distinct regimes coincide with a rapid and large change of resistance (and strain) when the pores collapse\cite{pruvost} followed by a slower decrease in resistance after the air is pushed out. It is expected that the linear approximation between the datasets of -10 kPa and -60 kPa does approximate the -40 kPa dataset well but not the -20 kPa. This result implies that the collapse/buckling of the structure happens between -10 and -40 kPa, which is expected to not be captured well by the linear model.

The overall identification results are summed up in Figure \ref{fig:curvaturereconstruction}(d). On average all the models capture the dynamics quite well. It can be observed that refining the WH model reduces the fitting error by 21.4\% (i.e. the NRMSE error decreased from 19.6 to 15.4\%). In addition, the WH model seems better in general than the Wiener and HS models, as seen in the average NRMSE fit. The Wiener and HS models also have a larger standard error (3.2 and 2.7\%) than the WH model (1.6\%), which implies a more consistent estimation result for the WH albeit both are better than the linear and HS model.  The RMS error (normalized by the maximum curvature) of the models was computed to be: $5.2 \pm 1.4, 6.5 \pm 1.9, 7.0 \pm 2.7$, and $10.42 \pm 7.98$\% for the WH, Wiener, HS, and linear models, respectively. These values again indicate that the nonlinear models are better estimators. The effect of these RMS errors scaled to the maximum curvature is shown in Figure \ref{fig:curvaturereconstruction}(e), which shows that the error reduction of the nonlinear model improves the curvature estimate significantly. 

The good NRMSE fits and low RMS errors imply that WH and its variation can provide a reasonable estimation of the curvature. Whereas nonlinearities are necessary to compensate for the nonlinear relation between resistance change and curvature in general. The piezoresistive sensorized actuator's ability to measure its strain indicates that it (at least) matches works primarily focused on empirical behavior of foam-like porous sensors and sensorized actuators\cite{murali,pruvost,foamsense,bilent,somm2019expanding}.  Whereas the sensorized piezoresistive bellow (not porous) in \cite{zhou} showed comparable performance in terms of estimation errors (albeit for the force) with around 4\% (they did not report a value for position estimation) versus our 5.2\%. Similarly, in \cite{Saadeh} a 90\% fit was acquired for stress estimation but using a commercial force sensing resistor, which makes absolute comparisons difficult. However, similar to their result the Wiener/HS models provide much better results than those seen with linear models. In addition, our results indicate that the WH is a better model structure for strain estimation. However, they did not investigate such a model structure for strain estimation.  

\subsection{Contractor - Mechanical \& Sensitivity Programming} 

Contracting actuators are among the most popular actuators as linear motion can be used for many applications. Contractors can benefit from porosity gradients as the location and number of spacers can have a significant impact on aspects such as their peak force and contraction ratio \cite{vacuumbellows}. Within this work, the contractor shown in Figure \ref{fig:gradact}(f) was printed. These contractors were printed by stacking multiple levels of porosity which was printed in a cylindrical form with a 25 mm diameter and a total height of 50 mm. Firstly, a low-porosity base layer of 5 mm was printed. Subsequently, a high porosity section of 40 mm height was printed. However, at heights of 14 and 28 mm a low porosity ring was also printed with a width of 5.6 mm. Lastly, after printing up to a height of 40 mm a low porosity top was printed of 5 mm height. The porous structure acted here as both a scaffold and a sensing structure. It allowed for printing a tall collapsible structure that allows air passage throughout its structure. Furthermore, the low porosity rings were printed to make a contraction the preferred deformation of the structure. By using low porosity rings and bottom/top caps, a path for air to travel through the entire structure was provided.

A set of contractors was investigated to see how the porosity magnitude in the high-porosity sections affected their mechanical properties and sensitivity. Specifically, contractors based on the coiling pattern at heights of 4, 6, 8, and 10 mm and a coil density $N$ of 3 (equivalent to porosities of 68, 76, 82, and 86\%) were printed. Whereas the top and end cap were made stiffer by using a low porosity (less than 5\%) and the rings were near-zero porosity. The experimental protocol was identical to that of the curvature sensing with the contractor hung in line with gravity and a single marker was added. The experiments were performed for three load cases: no-load, a 200-gram weight, and a dumbbell (500 grams).  

\begin{figure*}[h!] 
\centering
\includegraphics[width=0.9\textwidth]{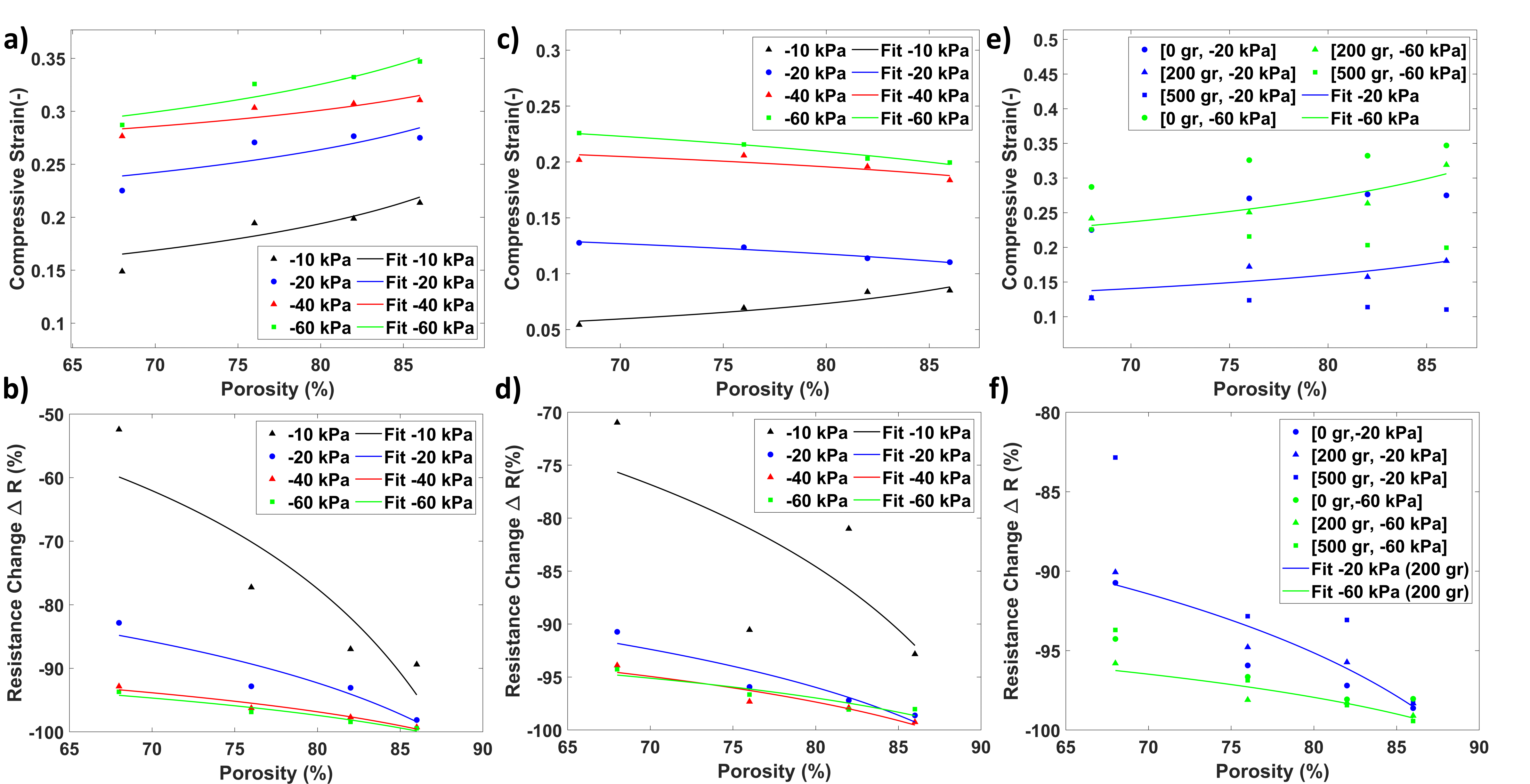}
\caption{a-f) The contracting actuator’s strain/resistance change versus porosity with power-law fits. a-d) The (a,c) strain and (b,d) resistance change versus porosity for the (a,b) no-load and (c,d) 500 gr load case for multiple levels of pressure, e,f) The (e) strain and (f) resistance change for multiple loads.}
\label{fig:cntrctr}
\end{figure*}

The averaged strain and resistance change for all contractors is shown in Figure \ref{fig:cntrctr} for different load cases. Within the experiments the maximum strain varied (on average) less than 10\% and was, therefore, not added to the graphs. In general, a similar trend can be seen for all data points for the no-load case with both the compressive strain and resistance change nonlinearly increasing with both the porosity and pressure. In contrast, the increase of a load negatively impacts the strain but the resistance change seems less affected. It can be seen that an increased load means that the resistance changes increased less with increasing porosity compared to the no-load. However, the strain becomes negatively impacted by increasing porosity for a 500 gr load whereas this decrease is less for a 200 gr load. This discrepancy implies that the relation of both the strain and resistance change is load-dependent. 

Higher porosity coincides, in general, with a higher change in resistance, which is to be expected as the final shape is smaller and the contact area increases more significantly. However, this increase does not correlate with a higher strain sensitivity for the 500 gr load. This contradiction is expected to be due to the load's effect on the contractor's stiffness. The addition of a load will increase the longitudinal stiffness as a constant force spring but not the radial stiffness. At a certain load, the radial stiffness will be lower than the longitudinal. At that point, the collapse will be more radially. Thereby still reducing the resistance significantly but reducing the strain.  Such change is more apparent for the higher porosity (softer) actuators as these have a lower radial stiffness overall. 

Interestingly, the trends of both the change in resistance and strain can be approximated by a power-law $p_f = C \left( 1-\phi/100 \right)^n$ for all load and pressure cases. Within this equation, variables $p_f, \phi, C, n$ represent the property of interest of the porous structure, the porosity (\%), and two fitting parameters. It has been shown that this power law can estimate changes in mechanical properties (yield stress, density, elastic modulus) of cellular solids (such as foams) \cite{gibson} and mechanical programming of soft actuators \cite{infoam}. Similarly, the results of Figure \ref{fig:cntrctr}(a-f) show that this empirical law approximates the curve quite well (the fitted parameters and $R^2$-values are shown in the Supporting Information (Table S1). The strain change for different loads is approximated quite well by a power-law fit with $R^2$-values on an average of 0.78 (all loads). Similarly, the change in resistance also fitted this empirical law quite well with $R^2$-values of on average 0.85 (all loads). However there is one exception and that is the -10 kPa pressure, which had only 51\% for the 500 gr load. One possible explanation could be that the initial contact resistance dominates as the collapse at this point was not sufficient for the lower porosity contractors. Or the outlier at 76\% porosity could have negatively impacted the fit due to contact resistance. In general, the relationship between porosity and resistance change is similar to the result of \cite{bilent}, which also showed increased sensitivity to pressure with increasing porosity (for a piezocapacitive sensor). These results provide evidence that the porosity can be used as a tool to program both the strain but also its sensitivity (i.e. resistance change). This power-law behavior with porosity was also seen for the behavior of a bending actuator in our earlier work \cite{infoam}. Although for low vacuum pressures (i.e. $\leq$-20 kPa) this approximation seems less correct.

In terms of sensitivity, this result indicates that the piezoresistive sensing approach can achieve very high sensitivity. Comparing it to the twisted-coiled polymer actuator of \cite{TCP} their resistance change was around 28\% for a strain of 12.5\% whereas the contractors have resistance changes of 90+\% with strains of around 20-30\%. Thereby indicating that if the hysteresis can be compensated a high sensitivity can be attained with these sensorized actuators. This comparison is also favorable when comparing with the inductive sensing for proprioceptive foam proposed in \cite{joe2021sensing}, which had inductance changes of 1 (\% normalized inductance)/ (strain \%) for strain sensing (i.e. for 30\% strain it would have 30\% change of inductance). 

\subsection{Reconstructing Contractor Deformation} 

To further validate the WH model's ability to provide good estimation over a wide range of porosity and load cases, the contractors were also investigated for their strain estimation performance. Both the highest and lowest porosity contractors with no load were also subjected to identification. In addition, the low porosity contractor was also identified for the 500-gram load to investigate the effect of the load on estimation performance. 

\begin{figure*}[h] 
\centering
\includegraphics[width=0.95\textwidth]{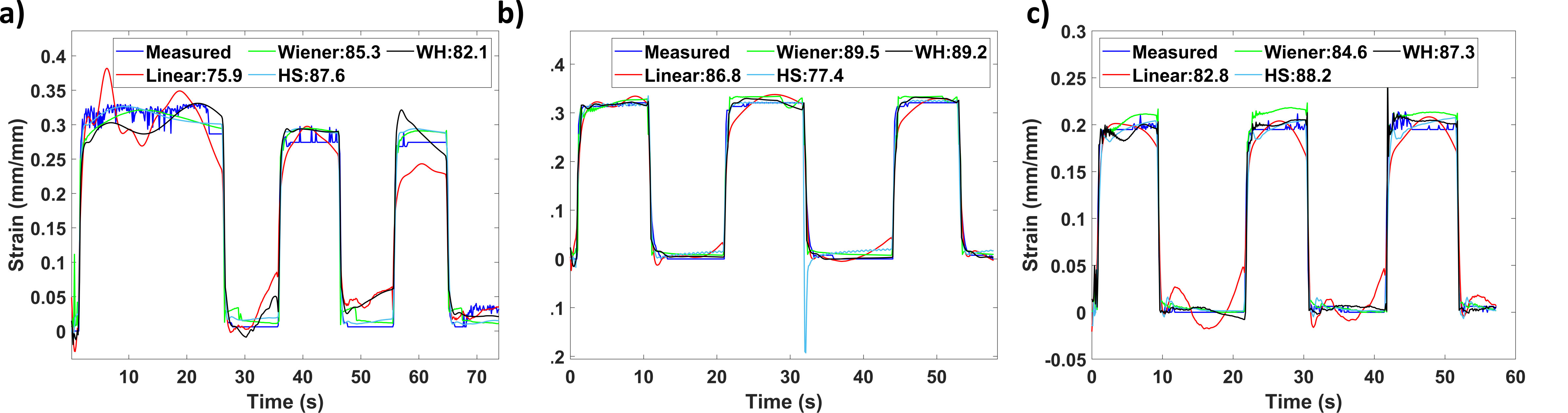}
\caption{The measured and estimated contraction by the identified models for the contractor system with the NRMSE fit in the legend. a-c) The results for the -40 kPa validation datasets for (a) the low porosity contractor (68\%)  with no load,  (b) the high porosity (82\%) contractor with no load, and (c) the low porosity contractor (68\%) with 500 gr load. Abbreviations: WH and HS mean Wiener-Hammerstein and Hammerstein models, respectively.}
\label{fig:defreconstruction}
\end{figure*}

All models were identified separately for the no load and 500 grams. This change was done based on the results of Figure \ref{fig:cntrctr}, which indicate that the relation between strain and resistance change is affected by the load. To investigate the effect of load changes, the no-load WH model was used as a filter to identify an additional linear transfer function that represents the additional load to evaluate whether that would improve estimation results. This identification approach gave the results of  Figure \ref{fig:defreconstruction}(a-c) for the -40 kPa validation dataset, which indicates overall good NRSME fits for all models. The lines in Figure \ref{fig:defreconstruction}(a-c) seem to approximate the real trajectory reasonably well they do not track them perfectly, which is especially notable in Figure \ref{fig:defreconstruction}(a) where qualitatively the most discrepancy is present. In general, the WH model performs the most consistently but the Wiener and HS models seem to be able to achieve similar performance for this dataset. Moreover, in Figure \ref{fig:defreconstruction}(a) the improvement for the Wiener model is very clear for the third pulse. This improvement could imply it is a better model type, but based on the other two an overfitting of the model on the identification dataset seems more likely. The additional fitting parameters of the linear transfer function after the Wiener model could lead to overfitting.

\begin{figure*}[h] 
\centering
\includegraphics[width=0.7\textwidth]{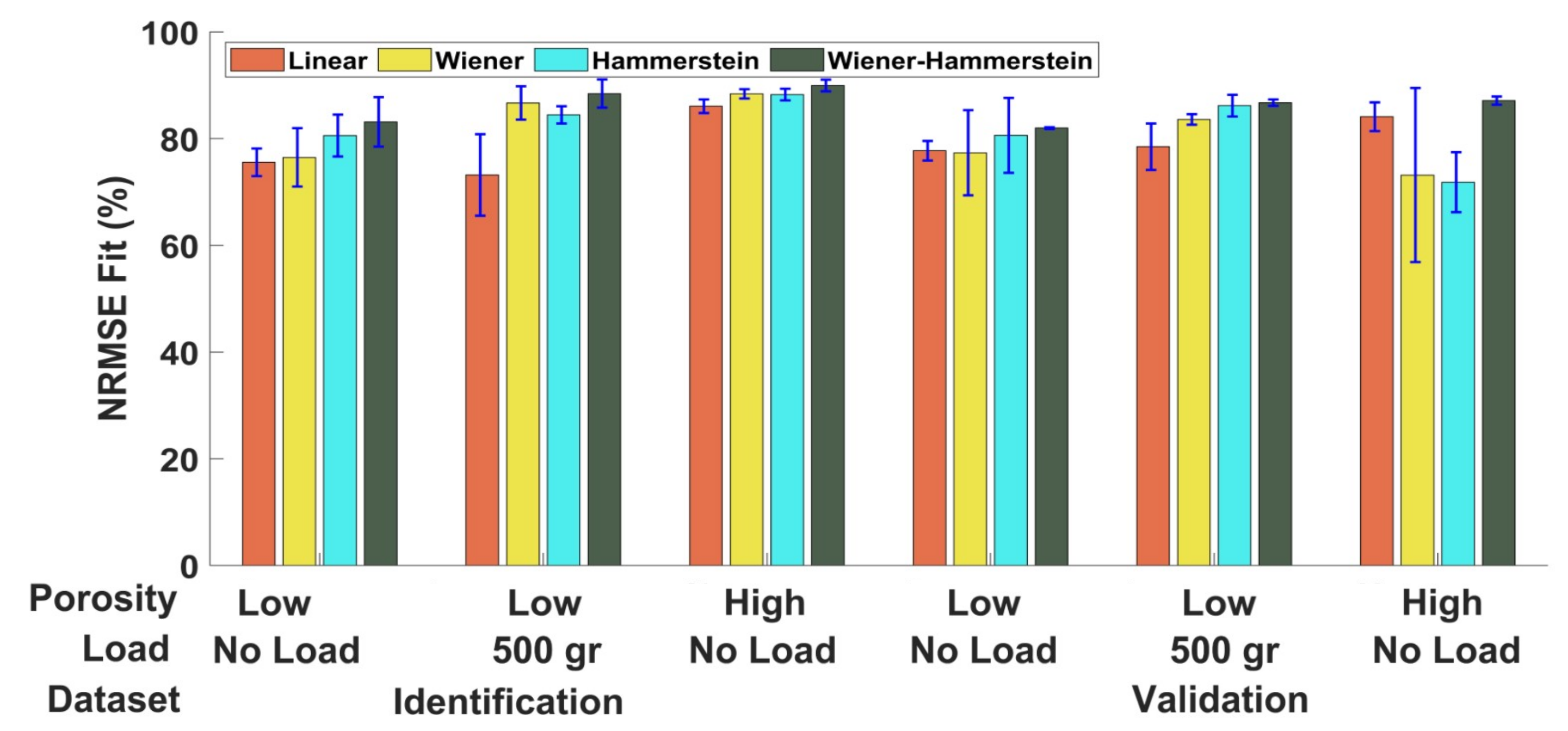}
\caption{Bar graph with NRMSE fits for different datasets and with standard errors indicated.}
\label{fig:defreconstruction2}
\end{figure*}

The results of the average NRMSE fit are shown in Figure \ref{fig:defreconstruction2}. In general, the NRMSE fits of the WH model outperformed the others on average. It can be noted that the others were still good (in most cases) but the Hammerstein and Wiener models performed better than the linear model except for one dataset. In general, the WH model is the superior option for all situations albeit with more improvement for certain datasets than others (as also seen in Figure \ref{fig:defreconstruction}). Especially, its low standard error over all datasets indicates good generalizations whereas other model types have some outliers. These results support our hypothesis that the WH model is a good representation of the structure of the underlying dynamics as it consistently achieves the highest accuracy. 

Similarly, the RMS error (normalized by the maximum strain) was evaluated. Averaged over all three datasets this gave [8.7, 7.1, 7.3, 5.8]\% for the linear, HS, Wiener, and WH models, respectively. These values indicate an overall good estimation of the models with the nonlinear models performing better and WH being the best overall. The effect on estimation is visualized in Supporting Video 2. 

To evaluate the consistency when the load changes the best WH model was also evaluated for the 500 gr load case. This model saw a decrease in the NRMSE fit to 61.73\% on average, which is a significant decrease from 82.6\% for the no-load case (averaged over all datasets). Similarly, the RMS error increased to 14.39\%, which is tripled compared to the no-load case. However, adding a transfer function in series (to represent the added mass) increased the NRMSE fit to 74.07\% and reduced the RMS error to 11.3\%. But these are worse than a linear model estimated to just the dataset itself making the model inferior. These results imply that although the model can capture the behavior for the same load, changes in the mechanical properties can severely impact the estimation performance. Such behavior could be mitigated by designing sensors only sensitive to one direction, or multiple smaller sensors with data fusion to improve estimation.

Lastly, it can be observed that this data supports the discussion of the previous section, by further providing evidence that the WH model is an appropriate structure for strain estimation with piezoresistive sensors. Furthermore, it shows that the WH model approach generalizes over multiple levels of porosity leading to models with similar magnitude of accuracy. 

\subsection{Three Degrees of Freedom Bending Segment} 
A three-degree of freedom (3DoF) bending segment provides motion in three directions (two rotating and compression), which can be used as a segment of a continuum arm. This 3DoF bending segment was investigated to validate our combination of the InFoam method and data-driven approach for a system with multiple inputs and outputs. The 3DoF bending segment was realized by placing three contractors in parallel (Figure \ref{fig:gradact}(f)). Three contractors were printed with the same dimensions as indicated in the previous section with a porosity of 76\%. The contractors were fixed using two rigid (PLA) 3D printed frames, which were designed as an equilateral triangle with lengths of 52 mm. Whereas the spacers (behind the black marker) have a height of 15 mm and the stage on top is an equilateral triangle (length of 20 mm) with rounded edges.

 This system differs from the previous two as it needs to combine the three outputs of the individual sensors to reconstruct the 3D deformation. Within this work, the models for this actuator were realized using three multi-input, single-output (MISO) systems in parallel. Every MISO model used the three changes of resistance as an input and one of the three deformations as an output.

For the experiments, a setup with three vacuum inputs and three resistance measurements was used. This setup was similar to both the bending and contracting actuator with a marker added for tracking the compression. In addition, an inertial measurement unit (IMU) was mounted on top to measure the angle in $x$ and $y$. The axis definition of this IMU is shown in  Figure \ref{fig:gradact}(f).

As the pressure input, both sequential and parallel activation patterns of the contractors were used. Either one, two, or three contractors were activated using the same on/off cycling as for the other one. In addition, the activation of two contractors at different levels of vacuum pressure was included as well. 

\begin{figure*}[h] 
\centering
\includegraphics[width=0.94\textwidth]{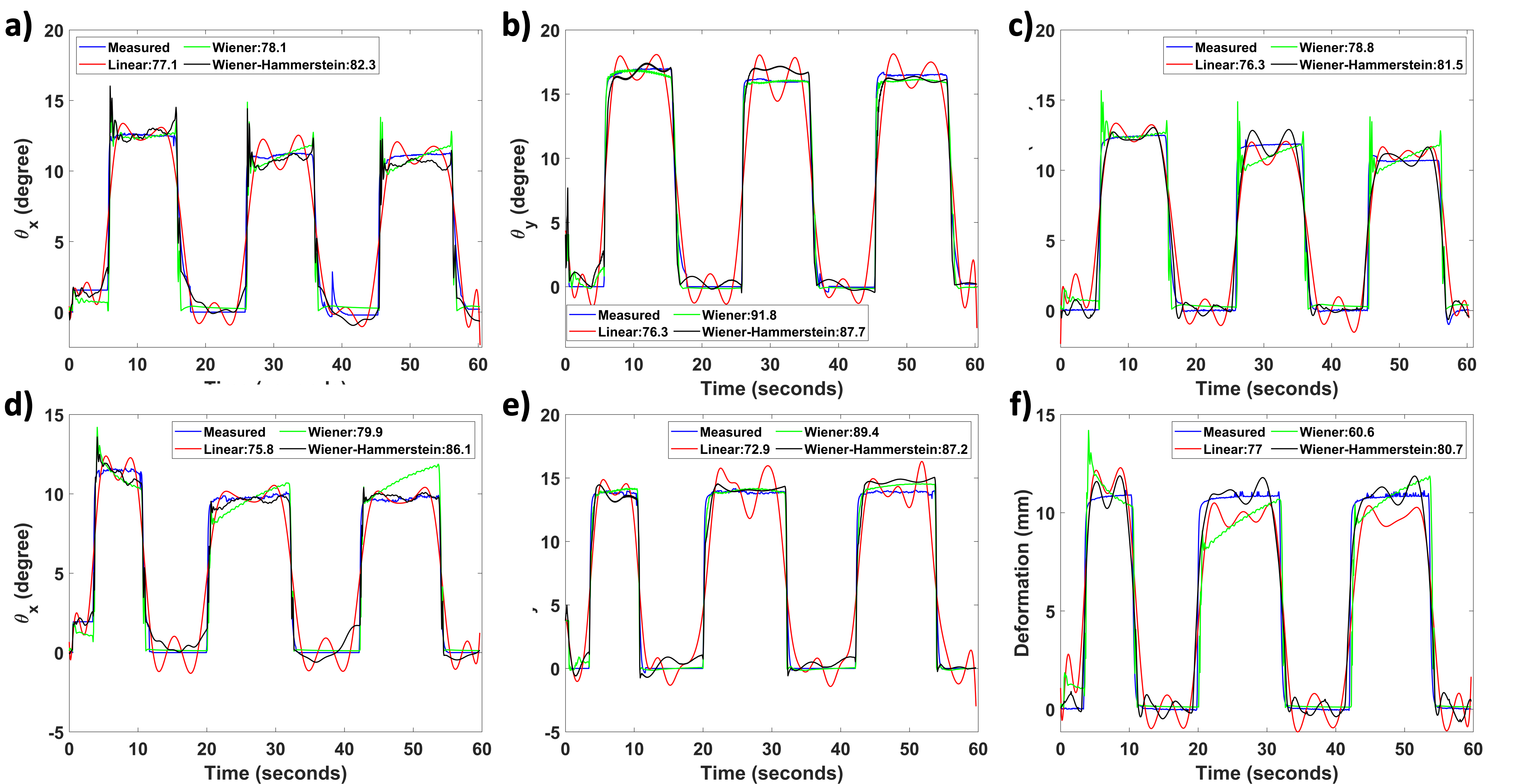}
\caption{ a-f) The 3DoF bending segment’s measured and estimated deformation for the angle in the (a,b) $x$ and (c,d) $y$-direction and (e,f) deformation in $z$  for the a,c,e) identification and b,d,f) validation dataset with their NRMSE fits included in the legend. Abbreviation: WH means Wiener-Hammerstein.}
\label{fig:defreconstruction_3c}
\end{figure*}

The realized 3DoF bending segment could bend ($\theta_x$,$\theta_y$) for 26.3/22.6 degrees. Whereas it had a maximum $\Delta z$ of 16 mm (axis definition in Figure \ref{fig:gradact}(f). Three individual identifications (one for each deformation) were performed with all three resistance changes as an input and one of the deformations as the output ($\theta_x,\theta_y,$ or $\Delta z$). The predictions of the identified models are seen in Figure \ref{fig:defreconstruction_3c}(a-f). The HS models had very poor NRMSE fits of below 70\% and were omitted. In general, the linear and WH models seem robust, as they stay around similar levels for all cases. In contrast, the Wiener model dips below 60\% and for some datasets (not shown) below 40\%. However, for steady-state situations, the WH model and Wiener models perform better than the linear model. This discrepancy is especially visible in the oscillations that the linear models predict during steady state. In general, the overall behavior seems better captured by the WH model. 

The NRMSE fit was computed by averaging the identification and validation datasets NRSME fits, which are shown in Table \ref{table:3dof}. It can be noted that the NRMSE fit is better for the WH model but not the Wiener model. A possible reason for this discrepancy is that the actuator is nonlinearly related to the contraction of an individual contractor \cite{Allen}, which might not be captured well by a Wiener model.  In contrast, the absence of nonlinearities could make the linear model more robust leading to better overall performance than the Wiener model. Similarly, the RMSE (normalized by maximum bending/deformation within each experiment) is 7.4$\pm$0.9/6.1$\pm$1.9 and 8.4$\pm$1.7 \% for the WH whereas the linear model was 10.1$\pm$0.8/9.5$\pm$1.1 and 9.9$\pm$1.0 \%. Implying again that the WH model is a good option for these systems. The estimation behavior is visualized in Supporting Movie 2.

\begin{table}[h!]
\caption{NRMSE fits (\%) for the 3DoF bending segment actuator for both angles (in $x$ and $y$) and translation in $z$.}
\begin{center}
\begin{tabular}{|c|| c | c| c |} 
 \hline
 \textbf{Type} & $\theta_x$ & $\theta_y$ & $\Delta$z \\  
 \hline\hline
 Linear & 76.0$\pm$1.5 & 77.44$\pm$2.6 &  76.38$\pm$1.95 \\ 
 \hline
 Wiener & 76.7$\pm$6.8 &  80.4$\pm$14.9 &  51.53$\pm$21.4\\
 \hline
 WH & 82$\pm$2.3 & 79.4$\pm$9.8 &   80.6$\pm$2.8 \\
 \hline
\end{tabular}
\label{table:3dof}
\end{center}
\end{table}

In general, these results support the hypothesis that the strain of the 3D-printed sensorized actuators can be estimated with identified models. Furthermore, the multi-input and output nature of this system indicates that the results of the bending actuator and contractor also hold for actuators with more degrees of freedom. Thereby supporting our hypothesis that the WH model is an appropriate model structure for piezoresistive sensing. In addition, it goes beyond the works discussed previously, which focused on single input-output systems, and shows that our approach generalizes to more complex strain estimation problems. Interestingly, the estimation performance stays similar in terms of magnitude for all three examined actuators but only for the WH and the linear models. Whereas both the Wiener and HS models underperform for this final system. Overall, the consistency of these results implies that the WH model is a reasonable approximation and could be a useful tool for strain estimation with good accuracy.

\subsection{Identified Model Properties}
The models that were identified were only discussed from their NRMSE fits in the previous sections. Other properties are interesting to discuss as well such as the number of poles and zeros (listed in the Supporting Information (Table S2)). For the linear models, these tend to be near the upper limit that was considered. However, their generalization was (comparatively) poor. Such that it seems likely that increasing the number of poles/zeros would not lead to improved model performance. In contrast, both the Wiener and WH are significantly below the upper limit but there does not seem to be a clear relationship/optimum. Thus, it also seems unlikely that increasing the number of poles/zeros would improve their fit as well. It can also be noted that the WH models are in general more computationally expensive. This property could be unfavorable for real-time problems such that the lower fits of the Wiener model could still make them a better candidate for closed-loop control.

It was hypothesized that the nonlinear function $g(.)$ should relate to the stress in some way. The estimated functions are shown in the Supporting Information for all the considered systems (Figures S3 and S4). The piecewise linear function’s estimated breakpoints (i.e. where the slope changes) were found to all fit an exponential function ($a e^{bx}$) with high $R^2$-values. This correlation between different actuators and the simple shape of the curve gives some support to the hypothesis that a simple nonlinear function relates the piezoresistivity and strain to a high accuracy. Interestingly, the shape described by our breakpoint ($a e^{bx}$) is similar to that of \cite{Saadeh}.  However, they estimated Wiener models for stress estimation with a Force Sensing Resistor with their $b$-values having an inverted sign in comparison. Such a change is equivalent to inverting the function that our models have. The similarity in shape between the curves could be interesting to further explore as it could lead to a useful yet simple relationship to relate both the stress and strain to the piezoresistivity. 

\section{Conclusion}
Sensor-integrated soft actuators have the potential to be used in a broad range of applications as they inherently provide proprioceptive data. By providing proprioceptive data these actuators can sense their deformation, which can be useful for applications such as strain estimation and feedback control. By realizing this proprioceptive ability through 3D printing, the fabrication of (geometrically) complex actuators that can close the loop themselves becomes possible.

Within this work, we have demonstrated that the InFoam method provides a simple approach to fabricating porous structures that are lightweight, provide air passage, and can be mechanically programmed while also integrating sensing capabilities. Thereby providing a versatile manufacturing approach for soft sensorized vacuum actuators. Furthermore, combining these printed actuators with a data-driven approach enables these actuators to estimate their deformation through an identified nonlinear model. These two aspects combined lead to a manufacturing process that provides tools for both fabrication (InFoam) and calibration for actuator usage (strain estimation through system identification).

The contractor results show that the InFoam method can set the porosity to tailor both the actuation and sensing behavior. Furthermore, both the strain and resistance change seem to behave in a (nearly) power-law relationship with the porosity (at the same pressure). Thereby providing evidence that porosity can be a versatile programming tool for programming both mechanical behavior and sensing, which is an extension of our previous results that showed it for mechanical properties only. However, our results indicate that both the strain and resistance change are affected by the load. Thus, further investigations are needed on how to mitigate/solve this issue to reduce the effect of the load on sensing behavior to make the strain estimation load independent. Although identifying an additional linear transfer function in series with the old Wiener-Hammerstein (WH) model reduced the error it was still inferior to an identified model for that specific load. 

In general, the identified WH model could accurately estimate their strain over time (on average NRMSE fits $\approx$83\% and relative RMS errors $\approx$6.6\%), which was significantly better than linear models in most cases. The results in this paper indicate that the WH model seems to be a versatile model structure over a broad range of porosities, types of deformation (curvature and translation), and multiple degrees of freedom (3DoF bending segment). Our data shows that the WH model could compensate for the nonlinearities and hysteresis of the sensorized actuator. It is expected that other piezoresistive sensors could also use this model structure, as it correlates with the underlying physics. These results indicate that combining system identification with piezoresistive sensing can be useful in estimating the porous actuator’s deformation, which is still a challenging problem due to the hysteresis. Therefore, the approach explored in this work can be useful for designers to realize soft robotics systems with integrated soft sensors that can compensate for the hysteresis inherent to piezoresistive sensors. Further research includes usage for closed-loop position control and methods to compensate for changes in dynamics. The latter was exemplified by the contracting actuator that the models were affected by changing loads. 

To conclude, the results in this paper indicate that the InFoam method is a versatile (yet simple) manufacturing method that can program both the mechanical behavior and sensing of soft (fluidic) actuators. Furthermore, it can realize proprioceptive actuators, which when combined with data-driven techniques can achieve decent strain estimation. Further research in this approach for both aspects is required to realize capabilities such as force estimation and multi-electrode sensing. The integration of multiple or specialized sensors in the actuator body could make sensing less dependent on the load and/or add spatial features possibly allowing for compensating changes in the load. Furthermore, the actuators needed a post-printing assembly step, removing this step is essential to realize direct 3D-printed sensorized actuators. However, both the multi-electrode array and full actuator printing will require the usage of multi-material printing for smart and soft materials, which is an important next step toward integrated manufacturing. 

\section{Experimental Section}

\subsubsection*{3D Printing of Soft Sensorized Vacuum Actuators}
The bending, contracting, and 3DoF bending segment actuators (Figure \ref{fig:gradact})(f)) were fabricated using a modified Creality Ender 5 Plus (Shenzhen Creality 3D Technology Co., Ltd., China). This modified printer incorporated a screw extruder to enable the usage of thermoplastic elastomers. For this work, we used TC7OEX-BLCK pellets (Kraiburg TPE, Germany) with a Shore Hardness of 70A and a volume resistivity of 10 $\Omega$cm, which were printed at 195$^{\circ}$C with a 0.6 mm nozzle. For this material, the coiling radius $R_c$ and height $H$ were determined to fit a linear function as $R_c=0.40H-0.3$ (range 2.5 to 10 mm) based on the methodology used in our earlier work \cite{infoam}. The $R_c$ results are shown graphically in Figure S1 including the fit. In addition, to compute the porosity ($\phi$ in \%) the bulk density ($\rho_b=$ 0.97 g/cm$^3$) was used in conjunction with the measured weight ($m$ (g)) and volume of printed cubes ($V$ (cm$^3$)) using $\phi = 100\left(1-\frac{m}{V \rho_b}\right)$ (rounded to the nearest whole number). All printed actuators were put into 0.4 mm thick heat-sealed styrene-ethylene-butylene-styrene (SEBS) sleeves and their respective PLA holders to finalize the actuators.

\subsubsection*{Data Acquisition during Identification Experiments}
The resistance change was measured using an Arduino Uno (Arduino AG, Italy) through a voltage divider and sent over serial to MATLAB (The Mathworks, USA). This voltage divider consisted of the sensor in series with a bias ($R_b$) resistor of 1k$\Omega$. The analog-to-digital conversion (ADC) was done with a 16-bit ADC, namely the ADS1115 (Texas Instruments, USA) on a breakout board (Adafruit Industries, USA). To track the deformation a webcam was used to capture images of the actuator. Black stripes were added to the sleeve (bending actuator) and the curvature over time was reconstructed by MATLAB using the stripes to fit a circle and compute the curvature. Similarly, a black marker was added to the contractor and 3DoF bending segment to track the compression.

For the 3DoF Bending segment the orientation was measured using an IMU (a 9DOF absolute orientation BNO055 (Bosch Sensortec, Germany)) on a breakout board (Adafruit Industries, USA) with the axis definition shown in Figure \ref{fig:defreconstruction_3c}(b). In this experiment, both the IMU and ADS1115 were connected to an Arduino Uno, which again send the data over serial to MATLAB. 

\subsubsection*{Identification Protocol - Bending Actuator and Contractor}
For identification, the actuators were actuated by four levels of vacuum (gauge pressure of -10, -20, -40, and -60 kPa). The pressure was applied in on/off pulses of ten seconds for three cycles (repeated in duplicate). In addition, a second experiment was conducted with a gradual increase of gauge pressure from -10 down to -60 kPa (with a ten-second hold at -20, -40, and -60 kPa), after which the pressure was increased up to -10 kPa (with ten second holds at the same pressures). This experiment was repeated for three different loads for the contractors with no load, a 200-gram weight, and a 500-gram dumbbell. 

When estimating the model parameters, all nonlinear models used piece-wise linear functions as the nonlinear functions and were explored in the range of five to ten breakpoints. The choice was made to go for ten piece-wise linear functions as these got good results during initial manual estimation. Other functions showed lower NRMSE fits during the initial manual estimation and were not considered after that stage. The optimal selection of the number of poles and zeros was obtained by estimating all models up to ten poles and ten zeros with the condition that the number of poles is bigger than the number of zeros (for causality). All models were estimated using MATLAB's System Identification Toolbox. Specifically, the provided linear transfer function(tfest) and Hammerstein-Wiener (idnlhw) estimation functions were used. The termination conditions were set to 200 iterations or when a local minimum was detected. The input delay was set to zero for both model types and all inputs were normalized to a fraction by dividing by 100. For the nonlinear models, two tries were done to minimize the odds of encountering an unfavorable initial condition. The breakpoints of the piecewise linear function were estimated by the idnlhw-function as well. The Wiener-Hammerstein model was estimated by first estimating a linear or Wiener model and then using the output of these models to estimate a Hammerstein or linear model. These were then combined to make the full Wiener-Hammerstein model. The best of either approach was chosen as the final model. Five datasets were used for the estimation and validation of the models. Three datasets were used for estimation (ramp, -10 kPa, and -60 kPa), and the other two for validation. The best models per model type were selected based on the highest average NRMSE fits over both identification and validation.

\subsubsection*{Identification Protocol - Three Degree of Freedom Bending Segment}

For the three degrees of freedom bending segment, both sequential and parallel activation patterns of the individual contractors were used. Either one, two, or three contractors were activated at -10, -20, -40, and -60 kPa (gauge). Within these experiments, cycles of 10 seconds of pressure on/off three times were performed (repeated in duplicate). In another pattern, two contractors were activated with one at gauge pressures of [-10, -20, -30] and one at [-40, -60] kPa. The timings for the high and low-pressure contractors were (5,25,45)/ (15,35,55) (turn on/off) and (5,45)/(35,65), respectively. The output data from these experiments was used for the identification of three models, i.e. separately for each degree of freedom using all three contractors as an input. The same set of models and evaluation criteria were used for the bending and contracting actuators with the same poles and zeros set for every input.  However, the selection of identification and validation datasets was different in incorporating the effect of different pumps being on. The -10 and -60 kPa datasets were used for identification. Whereas the -20 and -40 kPa datasets were used for validation.    

\section*{Acknowledgements} \par 
This work was partially funded by the 4TU Dutch Soft Robotics program.

\medskip

\bibliographystyle{IEEEtran}
\bibliography{template}  

\end{document}